\pdfoutput=1

\documentclass[11pt]{article}

\usepackage[]{ACL2023}

\usepackage{times}
\usepackage{latexsym}
\usepackage{graphicx}

\usepackage[T1]{fontenc}

\usepackage[utf8]{inputenc}

\usepackage{microtype}

\usepackage{inconsolata}
\usepackage{multirow}
\usepackage{graphicx}
\usepackage{enumitem}
\usepackage{listings}
\usepackage[ruled]{algorithm2e}
\usepackage{amsmath,amsthm,amsfonts,amssymb,bm}

\title{UNIMO-3: Multi-granularity Interaction for Vision-Language Representation Learning}

\author{Hao Yang \\
  Harbin Institute of Technology \\
  \texttt{hyang@ir.hit.edu.cn} \\\And
  Can Gao \\
  Baidu Inc., Beijing, China \\
  \texttt{gaocan01@baidu.com } \\\And
  Liu Hao \\
  Baidu Inc., Beijing, China \\
  \texttt{liuhao24@baidu.com } \\
  \AND
  Xinyan Xiao \\
  Baidu Inc., Beijing, China \\
  \texttt{xiaoxinyan@baidu.com } \\\And
  Yanyan Zhao \\
  Harbin Institute of Technology \\
  \texttt{yyzhao@ir.hit.edu.cn } \\\And
  Bing Qin \\
  Harbin Institute of Technology \\
  \texttt{qinb@ir.hit.edu.cn } \\}

\begin{document}
\maketitle
\begin{abstract}
Vision-and-language (VL) pre-training, which aims to learn a general representation of image-text pairs that can be transferred to various vision-and-language tasks. Compared with modeling uni-modal data, the main challenge of the VL model is: how to learn the cross-modal interaction from multimodal data, especially the fine-grained interaction. Existing works have shown that fully transformer-based models that adopt attention mechanisms to learn in-layer cross-model interaction can demonstrate impressive performance on various cross-modal downstream tasks. However, they ignored that the semantic information of the different modals at the same layer was not uniform, which leads to the cross-modal interaction collapsing into a limited multi-modal semantic information interaction. In this work, we propose the UNIMO-3 model, which has the capacity to simultaneously learn the multimodal in-layer interaction and cross-layer interaction. UNIMO-3 model can establish effective connections between different layers in a cross-modal encoder, and adaptively capture the interaction between two modalities at different levels. The experimental results show that our model achieves state-of-the-art performance in various downstream tasks, and through ablation study can prove that effective cross-layer learning improves the model's ability of multimodal representation.
\end{abstract}

\section{Introduction}
Vision-and-language pre-training (VLP) aims to use large-scale image-text pair data to learn and simulate the human ability to understand the world through vision and language. VLP has been proven can achieve excellent performances on various VL downstream tasks, such as visual question answering (VQA) \cite{antol2015vqa,goyal2017making}, visual entailment\cite{xie2019visual} and image-text retrieval\cite{lin2014microsoft,plummer2015flickr30k}. 
\begin{figure}[ht]
    \centering
    \includegraphics[width=8cm]{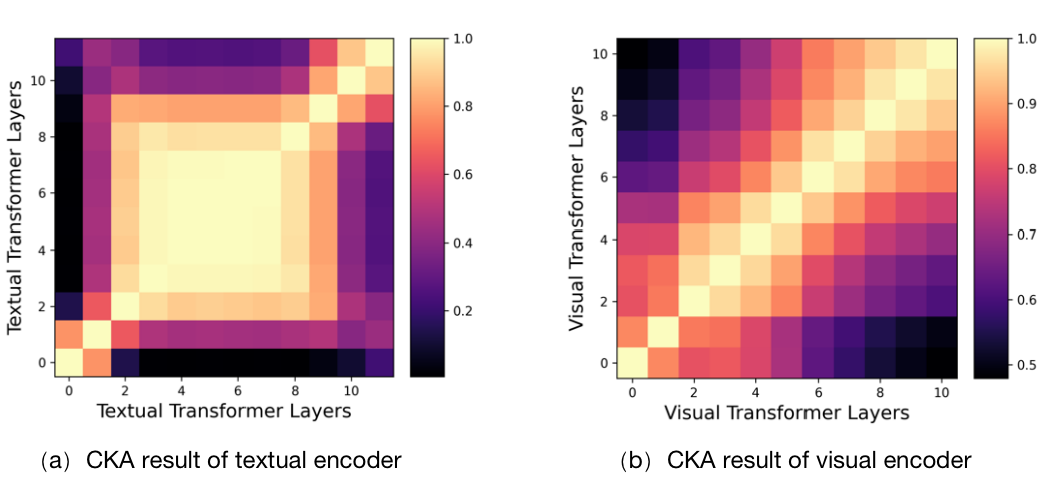}
    \caption{Centered kernel alignment (CKA) results of METER\cite{dou2022empirical} uni-modal encoders. }
    \label{fig-cka}
\end{figure}
Meanwhile, VLP can also benefit a series of uni-modal tasks\cite{wang2018glue,krizhevsky2009learning} without training a new model from scratch.

In other words, VLP aims to learn a shared embedding space for both text and images, the paired image-text samples are close to each other, while unpaired samples are separated from each other. It is relatively easy to learn such a shared embedding space for text-image matching, such as text-image retrieval task. Previous works \cite{radford2021learning,jia2021scaling} have achieved impressive results using image-text matching (ITM) pre-training task and contrastive learning techniques. However, for more complex tasks such as visual question answering (VQA), the VLP model needs to pay more attention to the fine-grained information in image-text pairs. How to improve the ability of the model to perceive, interact and fuse multimodal information with different granularity is crucial for the further development of the VLP model.

Previous works \cite{li2020unimo,tan2019lxmert,chen2020uniter,huang2020pixel,zhang2021vinvl} often rely on the object detection models to extract image features or encode the entire image, which dependent on external visual models (such as Fast(er) R-CNN \cite{ren2015faster}). \citet{zeng2021multi} proposed an explicit learning multi-granularity vision-language alignment method XVLM, but this method relies on the fine-grained annotation information in the data to establish fine-grained supervision. With the introduction of Vision Transformers \cite{dosovitskiy2020image}, the model using ViT as the visual encoder and combining with the Two-tower structure has achieved impressive performance in downstream tasks. These models \cite{dou2022empirical,wang2021simvlm} typically leverages the output of the final layer of the uni-modal encoder as input to the cross-modal encoder, the vision and language modalities can be jointly modeled by transformers. The output features of each layer of the uni-modal encoder exhibit noticeable differences in terms of information granularity. For instance, the output of lower layer text encoder contains word-level information, whereas the output of higher layer conveys global semantic information. Similar discrepancies can be observed in the image encoder. Therefore, these methods neglect to enhance the interaction modeling of multimodal information at different semantic granularity.

\citet{xu2022bridge} proposes to enhance the cross-modal alignment and fusion of uni-modal features at different semantic levels from bottom to top by establishing the bridge layer between the top layers of uni-modal encoders and the cross-modal encoder. We found that the output features from different layers in the end-to-end transformer-based model contain semantic information of different granularity. As shown in Figure ~\ref{fig-cka}, we adopt centered kernel alignment (CKA) to visualize the output layer features similarity of uni-modal encoder. Centered kernel alignment is a representation similarity metric that computes features normalized similarity in terms of the Hilbert-Schmidt Independence Criterion (HSIC). However, due to the modal differences, we found that the single link at the same layer of uni-modal encoder and multi-modal encoder is relatively limited. We propose the UNIMO-3 model which establishes effective links between the fusion encoder layer with all the uni-modal encoder layers, enabling the model to better leverage the multi-layer features of the uni-modal encoder and adaptively capture interactions between multimodal information at different granularities.

\begin{figure*}[ht]
    \centering
    \includegraphics[width=16cm]{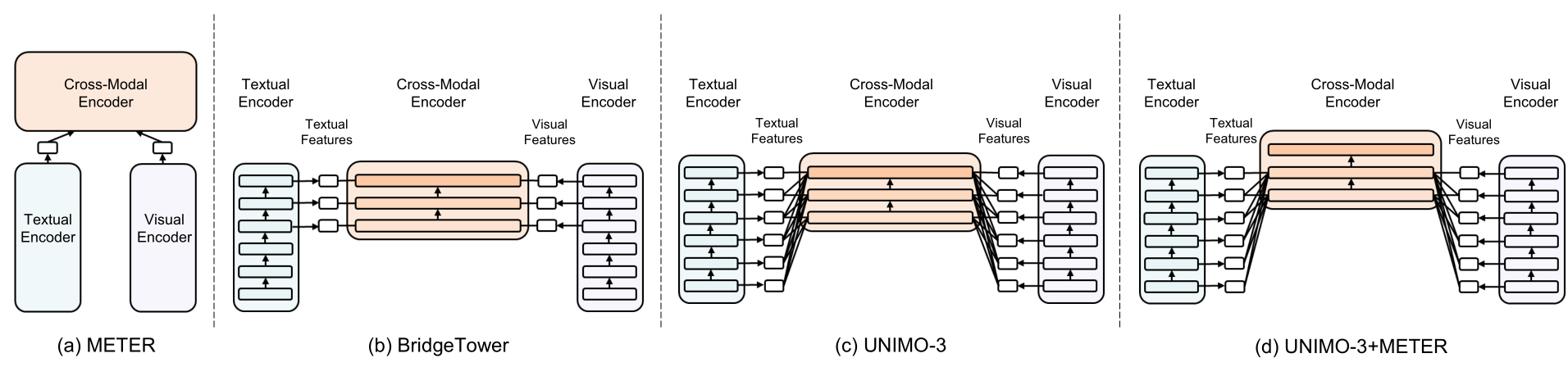}
    \caption{TWO-TOWER vision-language models: (a) METER\cite{dou2022empirical} model strcture. (b) BridgeTower\cite{xu2022bridge} structure, establish connections between top layers of the uni-modal encoder and the cross-modal encoder layers. (c) UNIMO-3 structure, cross-layer multi-granularity interaction. (d) UNIMO-3 and METER strcture, establish cross-layer interactive links only at the bottom layer, keeping the number of cross-modal encoder layers unchanged.  }
    \label{fig-model_small}
\end{figure*}

In our extensive set of experiments, we show that our method achieves competitive performance compare with the VLP models that with same pre-training data and a similar number of parameters. Specifically, with only 4M images for pre-training, UNIMO-3 achieves a new SOTA of 78.88\% on the VQAv2 test-std set. UNIMO-3 also outperform other baseline models when further scaling the model, achieves 81.27\% on the VQAv2 test-std set. This result outperforms even the models with more pre-training data and larger number of paramters.

Our main contributions are as follows: (1) We found that the existing VLP models lack to learn the cross-modal interaction in different granularity. (2) We introduce a new model structure, the UNIMO-3 model, which utilizes a fusion encoder that adaptively selects multi-layer output features from uni-modal encoders using the gating mechanism to enable multi-granularity interaction of multimodal information. (3) Through a series of experiments, UNIMO-3 achieves state-of-the-art results and outperforms other models on a range of downstream validation datasets.

\section{Related Work}
\subsection{Vision-Language Pre-training Models}
The VLP model aims to learn a shared embedding space for both text and images by utilizing large-scale image-text pair datasets that are collected from the public web. Paired images and texts are positioned in close proximity within the feature space, while unpaired samples are separated from each other. VLP model has been proven to perform well in downstream visual tasks, language tasks and multimodal tasks. Some early VLP models followed a pipeline approach and utilized external models as visual encoder such as Faster R-CNN and ResNet to extract visual features.
Since the emergence of Vision Transformer, some work has made better performance in downstream tasks by using ViT as a visual encoder and combining with the two-tower model structure. These models generally take the output feature of uni-modal encoder's last layer as the input of cross-modal fusion encoder, and rely on the cross-modal learning mechanism (such as dot product, co-attention mechanism) in the fusion encoder to learn the shared embedding space for both text and images. 

\citet{dou2022empirical} compared the effects of these models on different structural designs and the selection of pre-training tasks through rich experiments, and selected the optimal combination to achieve the state-of-the-art effect in multiple downstream tasks, as illustrated in Figure ~\ref{fig-model_small} (a). \citet{xu2022bridge} proposed to enhance the interaction of cross-modal information at different semantic levels by using the bridge layer to establish connections between top layers of the uni-modal encoder and the cross-modal encoder layers, which resulted in improved performance, as illustrated in Figure~\ref{fig-model_small} (b). As shown in Figure~\ref{fig-model_small} (c) and (d), we draw inspiration from the multi-layer feature utilization, we believe that the establishment of cross-layer connection rather than the interaction of the same layer can bring multi-granularity interaction of multi-modal information and achieve fine-grained cross-modal fusion.

\section{Approach}
\subsection{Visual Encoder}
Sinece the CLIP's visual encoder has been proven benificial for downstram VL tasks in previous works \cite{xu2022bridge,dou2022empirical,2021How}, we adopt CLIP-ViT-B/16 as the pre-trained visual encoder. Given the input image $I \in \mathbb{R}^{3 \times H \times W} $ where $3$, $H$ and $W$ represent the number of channels, height and width of the image, the ViT split the image into $N$-patch sequence,  $N = \frac{H \times W}{P^{2}}$ where $(P,P)$ is the image patch resolution and a patch $p \in \mathbb{R}^{3 \times p^{2}}$. The input visual representation:
\begin{gather}
V_{0} = [E_{[class]};p_{1}W^{p}_{1};...;p_{N}W^{p}_{N}] + V^{pos}, \tag{1}
\end{gather}
where $V_{0} \in \mathbb{R}^{D_{v} \times (N+1)}$, $E_{[class]}$ represent the prepended token to the pach sequence, $W^{p} \in \mathbb{R}^{D_{v} \times (3 \times p^{2})} $ is the trainable linear projection layer, $V^{pos} \in \mathbb{R}^{D_{v} \times (N+1)}$ is learnable position embeddings, $D_{v}$ is the dimension of the visual encoder. The visual representation in the $l$-th layer visual encoder:
\begin{gather}
V_{l} = Encoder_{l}^{V}(V_{l-1}), l=1,...,L_{V}, \tag{2}
\end{gather}
where $L_{V}$ is the number of visual encoder layers.

\subsection{Textual Encoder}
Similar to previous work, we adopt RoBERTa-base as our textual encoder. Given the $M$-word input sentence  $S = (w_{1},w_{2},...,w_{M})$, we first add "[<s>]" token and "[</s>]" token at the sequence S start position and end position, and then tokenize the obtained new sequence. The input textual representation:
\begin{gather}
T_{0} = [E_{[<s>]};E_{w_{1}};...;E_{w_{M}};E_{[</s>]}] + T^{pos}, \tag{3}
\end{gather}
where $T_{0} \in \mathbb{R}^{D_{t} \times (M+2)}$ is the word embedding matrix, $M$ is the number of tokens, $D_{t}$ is the dimension of the textual encoder, and $T^{pos}$ is the positional embeddings matrix. The textual representation in the $l$-th layer textual encoder:
\begin{gather}
T_{l} = Encoder_{l}^{T}(T_{l-1}), l=1,...,L_{T}, \tag{4}
\end{gather}
where $L_{T}$ is the number of textual encoder layers.

\begin{figure*}[ht]
    \centering
    \includegraphics[width=16cm]{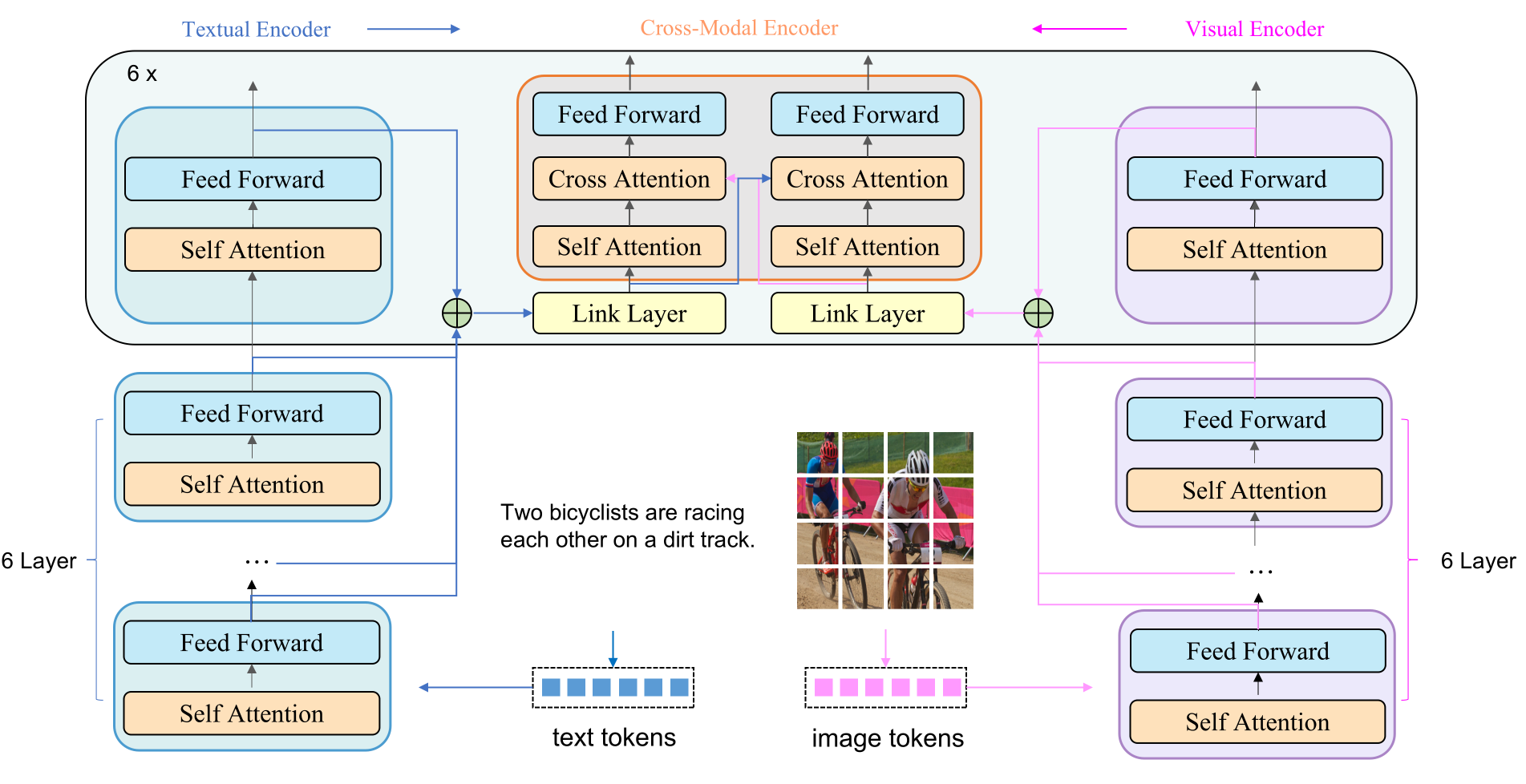}
    \caption{The overview of UNIMO-3 model architecture. }
    \label{fig-model}
\end{figure*}

\subsection{Cross-modal Encoder}
The fusion encoder plays a crucial role in the VLP model as it facilitates cross-modal interactive learning. Previous research has demonstrated that the transformer structure, which incorporates cross-attention mechanism and self-attention mechanism, yields optimal performance for cross-modal interactive learning in the fusion encoder. Our UNIMO-3 model is based on a similar structure, with each fusion encoder layer consisting of a visual and text component. These components comprise a multi-headed self-attention (MSA) block, a multi-headed cross-modal attention (MCA) block, and an FFN block. 
\begin{gather}
    Z^{T}_{l},Z^{V}_{l} = Encoder^{F}_{l}(\widetilde{Z}^{T}_{l-1},\widetilde{Z}^{V}_{l-1}), l=1,...,L_{F} \tag{5}
\end{gather}
where $Z^{T}_{l},Z^{V}_{l}\in\mathbb{R}^{D_{f}}$ represent the output features of the $l-th$ fusion encoder layer, and  $\widetilde{Z}^{T}_{l-1},\widetilde{Z}^{V}_{l-1}\in\mathbb{R}^{D_{f}}$ represent the input of the $l-th$ fusion encoder layer. $L_{F}$ is the number of fusion encoder layers.

Our proposed UNIMO-3 model utilizes a gate-based cross-layer feature selection method. The input features of the fusion encoder are derived from multi-layer output features of the uni-modal encoder rather than solely from the last layer. $Z^{T}_{0}$ and $Z^{V}_{0}$ are initialized with the last-layer representations from pre-trained uni-modal encoders before the start interaction layer $S$:
\begin{gather}
    Z^{T}_{0} = W^{T}T_{L_{S-1}} + T^{type}, \tag{6} \\
    Z^{V}_{0} = W^{V}V_{L_{S-1}} + V^{type}, \tag{7} 
\end{gather}
where $W^{T}\in\mathbb{R}^{D_{f} \times D_{t}}$, $W^{V}\in\mathbb{R}^{D_{f} \times D_{v}}$, $T^{type}$, $V^{type}$ are learnable parameters.

In each layer $l$ of the fusion encoder in UNIMO-3, the input involves interactions between the output of the previous layer $l$-1 and the output features of the uni-modal encoder across multiple layers, which is different from BridgeTower \cite{xu2022bridge} that only considering a single layer. We utilized gating mechanisms for selecting and filtering uni-modal feature across layers, which allows interactions between uni-modal information of different granularities in cross-modal interactions. Each textual parts in fusion encoder layers are connect with textual encoder layers: 
\begin{gather}
    g_{t} = Sigmoid(W^{G_{T}}T_{L_{t}}+b_{T}+Z^{T}_{l-1}), \tag{8} \\
    \widetilde{Z}^{T}_{l-1}=LayerNorm(\sum^{l-1}_{L_{t}}Z^{T}_{l-1}+g_{t}T_{L_{t}}), \tag{9}
\end{gather}
where $g_{t}$ represent the gate value, $L_{t} = 1,...,l-1$ is the textual encoder layer, $W^{G_{T}},b_{T}$ are learnable parameters. We adopt sigmoid function to calculate the gate value of the textual feature $T_{L_{t}}$ and link with the cross-modal layer base on the gate value. We follow BridgeTower to use (Add$\&$Norm) as the link layer. It is similar for visual parts:
\begin{gather}
    g_{v} = Sigmoid(W^{G_{V}}V_{L_{v}}+b_{V}+Z^{V}_{l-1}), \tag{10} \\
    \widetilde{Z}^{V}_{l-1}=LayerNorm(\sum^{l-1}_{L_{v}}Z^{V}_{l-1}+g_{v}T_{L_{v}}), \tag{11}
\end{gather}
This structural-level enhancement improves the model's ability to handle fine-grained multi-modal information.

\begin{table*}[]
\scalebox{0.9}{
\begin{tabular}{lrcccccc}
\multirow{2}{*}{Model} & \#Pre-train & \multicolumn{1}{c|}{Visual}            & Test-Dev & \multicolumn{4}{c}{Test-Standard} \\
                       & Images      & \multicolumn{1}{c|}{backbone}          & Overall  & Yes/No & Number & Other & Overall \\ \hline
\multicolumn{8}{l}{Base-Size models}                                                                                         \\ \hline
ViLT-base              & 4M          & \multicolumn{1}{c|}{ViT-B-384/32}      & 71.26    & -      & -      & -     & -       \\
UNITER                 & 4M          & \multicolumn{1}{c|}{Faster R-CNN}      & 72.70    & -      & -      & -     & 72.91   \\
VILLA                  & 4M          & \multicolumn{1}{c|}{Faster R-CNN}      & 73.59    & -      & -      & -     & 73.67   \\
UNIMO-2-base           & 4M          & \multicolumn{1}{c|}{ViT-B/16}          & 76.31    & -      & -      & -     & 76.42   \\
ALBEF-base*            & 14M         & \multicolumn{1}{c|}{DeiT-B-224/16}     & 75.84    & -      & -      & -     & 76.04   \\
UNIMO-2-base           & 4M          & \multicolumn{1}{c|}{ViT-B/16}          & 76.31    & -      & -      & -     & 76.42   \\
METER-CLIP-ViT-base    & 4M          & \multicolumn{1}{c|}{CLIP-ViT-B-224/16} & 77.68    & 92.49  & 58.07  & 69.2  & 77.64   \\
OFA-base*              & 54M         & \multicolumn{1}{c|}{ResNet-101}        & 77.98    & -      & -      & -     & 78.07   \\
SimVLM-base            & 1.8B        & \multicolumn{1}{c|}{ResNet-101}        & 77.87    & -      & -      & -     & 78.14   \\
BLIP-base*             & 129M        & \multicolumn{1}{c|}{DeiT-B-224/16}     & 78.24    & -      & -      & -     & 78.17   \\
Bridge-Tower-base      & 4M          & \multicolumn{1}{c|}{CLIP-ViT-B-224/16} & 78.66    & 92.92  & \textbf{60.69}  & 70.51 & 78.73   \\
UNIMO-3-base           & 4M          & \multicolumn{1}{c|}{CLIP-ViT-B-224/16} & \textbf{78.75}    & \textbf{93.20}  & 60.65  & \textbf{70.60} & \textbf{78.88}   \\
\hline
\multicolumn{8}{l}{Large-Size Models}                                                                                        \\ \hline
UNITER-Large           & 4M          & \multicolumn{1}{c|}{Faster R-CNN}      & 73.82    & -      & -      & -     & 74.02   \\
VILLA-Large            & 4M          & \multicolumn{1}{c|}{Faster R-CNN}      & 74.69    & -      & -      & -     & 74.87   \\
UNIMO-Large            & 4M          & \multicolumn{1}{c|}{Faster R-CNN}      & 75.06    & -      & -      & -     & 75.27   \\
VinVL-Large            & 5.7M        & \multicolumn{1}{c|}{ResNeXt-152}       & 76.52    & 92.04  & 61.50  & 66.68 & 76.63   \\
SimVLM-Large           & 1.8B        & \multicolumn{1}{c|}{ResNet-152}        & 79.32    & -      & -      & -     & 79.56   \\
VLMo-Large             & 4M          & \multicolumn{1}{c|}{BEiT-L-224/16}     & 79.94    & -      & -      & -     & 79.98   \\
OFA-Large              & 54M         & \multicolumn{1}{c|}{ResNet-152}        & 80.43    & 93.32  & \textbf{67.31}  & 72.71 & 80.67   \\
BridgeTower-Large      & 4M          & \multicolumn{1}{c|}{CLIP-ViT-B-224/14} & 81.25    & 94.69  & 64.58  & \textbf{73.16} & 81.15   \\

UNIMO-3-large      & 4M          & \multicolumn{1}{c|}{CLIP-ViT-B-224/14} & \textbf{81.26}    & \textbf{94.86}  & 65.12  & 73.02 & \textbf{81.27}   \\

\hline
\multicolumn{8}{l}{Huge or even Larger Size Models}                                                                          \\ \hline
METER-HUGE             & 14M         & \multicolumn{1}{c|}{Florence-CoSwin-H} & 80.33    & 94.25  & 64.37  & 72.30 & 80.54   \\
OFA-HUGE               & 54M         & \multicolumn{1}{c|}{ResNet-152}        & 82.00    & 94.66  & 71.44  & 73.35 & 81.98   \\
Flamingo               & 2.3B        & \multicolumn{1}{c|}{NFNet-F6}          & 82.00    & -      & -      & -     & 82.10   \\
CoCa                   & 4.8B        & \multicolumn{1}{c|}{ViT-G-288/18}      & 82.30    & 94.55  & 70.25  & 74.46 & 82.33   \\
BEiT-3                 & 28M         & \multicolumn{1}{c|}{BEiT-3}            & 84.19    & 96.43  & 73.63  & 75.92 & 84.18   \\
PaLI                   & 1.6B        & \multicolumn{1}{c|}{ViT-E-224}         & 84.30    & 96.13  & 69.07  & 77.58 & 84.34  
\end{tabular}
}
\caption{Comparisons with previous models on visual question answering (VQAv2). The best score is bolded. The models are divided into base size and large/huge size. B, N and M in ViT-B-N/M denote the model size, image resolution and patch size, respectively. * indicates that the model also uses VG-QA data to fine-tune on VQAv2. $\star$ denotes the model is trained from scratch. ``\# Pre-train Images'' denotes the number of images in VLP (the images for pre-trained visual and textual backbones are not counted).}
\label{result-vqa}
\end{table*}

\begin{table*}[]
\scalebox{0.95}{
\begin{tabular}{lccccccccc}
\multicolumn{1}{l|}{Model}               & \multicolumn{2}{c|}{SNLI-VE}                         & \multicolumn{7}{c}{Flickr30k}                                                                                         \\
\multicolumn{1}{l|}{}                    & dev            & \multicolumn{1}{c|}{test}           & IR@1           & IR@5           & IR@10          & TR@1           & TR@5           & TR@10           & RSUM           \\ \hline
\multicolumn{10}{l}{Pre-trained with \textgreater{}4M images}                                                                                                                                                           \\ \hline
\multicolumn{1}{l|}{ALIGN(1.8B)}         & -              & \multicolumn{1}{c|}{-}              & 84.90          & 97.40          & 98.60          & 95.30          & 99.80          & 100.0           & 576.0          \\
\multicolumn{1}{l|}{ALBEF(14M)}          & 80.80          & \multicolumn{1}{c|}{80.91}          & 85.60          & 97.50          & 98.90          & 95.90          & 99.80          & 100.0           & 577.7          \\ \hline
\multicolumn{10}{l}{Pre-trained with =4M images}                                                                                                                                                                        \\ \hline
\multicolumn{1}{l|}{UNIMO}          & 80.10          & \multicolumn{1}{c|}{79.10}          & 74.66          & 93.40          & -              & 89.70          & 98.40          & -               & -              \\
\multicolumn{1}{l|}{UNIMO-2}        & \textbf{81.97} & \multicolumn{1}{c|}{\textbf{81.48}} & 80.14          & 95.58          & -              & 92.01          & 99.31          & -               & -              \\
\multicolumn{1}{l|}{UNITER-Large}        & 79.39          & \multicolumn{1}{c|}{79.38}          & 75.60          & 94.10          & 96.80          & 87.30          & 98.0           & 99.20           & 550.9          \\
\multicolumn{1}{l|}{UNIMO-Large}         & 81.11          & \multicolumn{1}{c|}{80.63}          & 78.00          & 94.20          & 97.10          & 89.40          & 98.80          & 99.80           & 557.5          \\
\multicolumn{1}{l|}{ALBEF}               & 80.14          & \multicolumn{1}{c|}{80.30}          & 82.80          & 96.70          & 98.40          & 94.30          & 99.40          & 99.80           & 571.4          \\
\multicolumn{1}{l|}{METER-CLIP-ViT} & 80.86          & \multicolumn{1}{c|}{81.19}          & 82.20          & 96.30          & 98.40          & 94.30          & 99.60          & 99.90           & 570.7          \\
\multicolumn{1}{l|}{Bridge-Tower}        & 81.11          & \multicolumn{1}{c|}{81.19}          & \textbf{85.80} & \textbf{97.60} & 98.90          & 94.70          & 99.61          & \textbf{100.00} & \textbf{576.6} \\
\multicolumn{1}{l|}{UNIMO-3}             & 81.29          & \multicolumn{1}{c|}{81.23}          & 84.94          & 97.44          & \textbf{99.00} & \textbf{95.40} & \textbf{99.70} & 99.90           & 576.4     
\end{tabular}
}
\caption{Comparisons with models pre-trained with 4M images on visual entailment, and
image retrieval (IR) and text retrieval (TR) tasks.}
\label{result-flickr-snlive}
\end{table*}

\subsection{Pre-training Tasks}
 As observed in the ablation experiments of METER \cite{dou2022empirical}, in contrast to that the masked image modeling tasks lead to performance improvement in the pre-training of the region-based VLP models, the masked image modeling tasks resulted in a decrease in performance for transformer-based VLP model. Meanwhile, they found that the masked language modeling (MLM) task and image-text matching (ITM) task can bring performance improvements on downstream tasks. Therefore, we pre-train our UNIMO-3 model with MLM task and ITM task.

\begin{figure*}[ht]
    \centering
    \includegraphics[width=10cm]{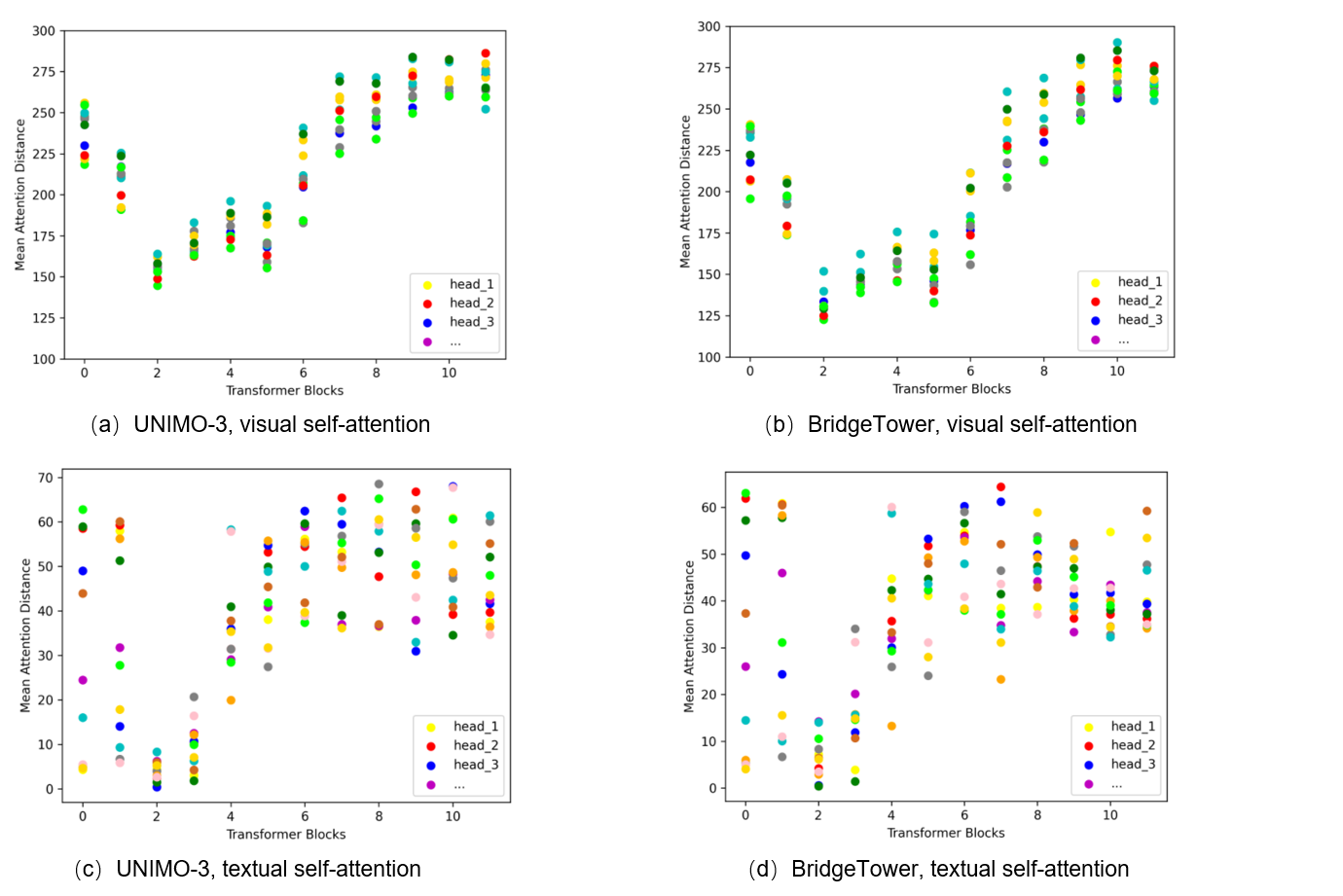}
    \caption{The average attention distance of BridgeTower and UNIMO-3.}
    \label{fig-visual}
\end{figure*}
 
 \textbf{Masked Language Modeling. }Masked Language Modeling (MLM) is widely applied in both language pre-training and vision-language pre-training and has poven to be useful. For the input image-text token sequence, we randomly samples and masks 15\% of tokens in the sequence while keeping the input image patch sequence untainted, similar to UNITER. The goal of masked language modeling task is to predict these masked tokens based on their surrounding context and visual information.

 \textbf{Image-Text Matching. }Image-Text Matching (ITM) aims to determine whether the input image and text are matched or mismatched and is widely used in previous works. The model is given a batch of matched or mismatched image-text pairs, we apply a binary classifier on the concatenated of final representaions of the cross-modal encoder to predict whether the image-text pair is matched or not.

\section{Experiment}
\subsection{Implementation Details}
Our pre-training data are composed of four existing image-text pairs datasets: COCO \cite{lin2014microsoft}, Visual Genome (VG) \cite{krishna2017visual}, Conceptual Captions (CC) \cite{sharma2018conceptual} and SBU Captions \cite{ordonez2011im2text}, which have also been widely used in previous VLP models and contain 4M images in total. We utilize CLIP-ViT-224/16 and RoBERTa-base to initialize the visual encoder and textual encoder of UNIMO-3. And the fusion encoder that consist of external cross-modal layer and internal cross-modal layer have a total of 6 layers, each with a hidden dimension of 768, intermediate size of feed-forward networks of 3, 072 and the number of heads of 12.
The maximum length of the text sequence is set as 50. The image size is set to $224\times224$ for pre-training. Same as previous works \cite{dou2022empirical,li2021align}, we apply RandAugment \cite{cubuk2020randaugment} for data augmentation and use the AdamW \cite{loshchilov2017decoupled} optimizer with a base learning rate of 1e-5 and weight decay of 0.01. The learning rate is warmed up for $10\%$ of the total training steps and then decayed linearly. Following previous works, the learning rate of the cross-modal encoder is five times higher than that of uni-modal encoders. We pre-train UNIMO-3 for 100k steps on 8 NVIDIA A100 GPUs with a batch size of 4, 096. 

We finetuning the UNIMO-3 model on the visual question answering (VQAv2) \cite{goyal2017making}, visual entailment (SNLI-VE) \cite{xie2019visual}, and image-text retrieval (Flickr30K) \cite{young2014image} tasks to evaluate the model's performance. Following BridgeTower, we set image resolution as $384\times384$ for these downstream tasks, except for VQAv2 as $576\times576$. And we also convert VQAv2 to a classification task with 3, 129 answer classes for fair comparison with previous works \cite{goyal2017making,teney2018tips}.

\begin{table*}[]
\scalebox{0.95}{
\begin{tabular}{l|cccccccc|l}
Model        & MNLI  & QQP   & QNLI  & SST-2 & CoLA  & STS-B & MRPC  & RTE   & \multicolumn{1}{c}{AVG} \\
             & 392k  & 363k  & 108k  & 67k   & 8.5k  & 5.7k  & 3.5k  & 2.5k  & \multicolumn{1}{c}{}    \\ \hline
METER PT   & 87.25 & 89.09 & 92.72 & 94.65 & 61.14 & 90.65 & 92.12 & 79.06 & 85.84  \\
BridgeTower PT      & 87.27 & 89.21 & 92.78 & 94.95 & 61.79 & 90.85 & 92.56 & 79.78 & 86.15 (+0.31)            \\
UNIMO-3 PT & \textbf{87.28} & \textbf{89.09} & \textbf{92.69} & \textbf{94.86} & \textbf{62.82} & \textbf{90.19} & \textbf{92.81} & \textbf{80.52} & \textbf{86.28 (+0.44)}            \\ \hline
RoBERTa-base & 87.55 & 89.22 & 92.87 & 95.07 & 63.21 & 90.70 & 92.89 & 79.90 & 86.43             
\end{tabular}
}
\caption{Fine-tuning performance of text encoders (RoBERTa$\rm _{BASE}$) on GLUE dev sets before and after VLP. PT is short for Pre-Training. We report average scores and standard deviations over three runs of different random seeds. Matthews correlations are reported for CoLA, F1 scores are
reported for QQP and MRPC, and Spearman correlations are reported for STS-B. The average of matched and mismatched
accuracy scores are reported for MNLI.}
\label{result-glue}
\end{table*}

\begin{table*}[]
 \centering
\begin{tabular}{l|cc|l}
Model                      & CIFAR-10 & CIFAR-100 & \multicolumn{1}{c}{AVG} \\ \hline
 METER PT       & 98.46    & 89.52     & 93.99          \\
 BridgeTower PT & 98.48    & 90.20     & 94.34 (+0.35)           \\
 UNIMO-3 PT     & \textbf{98.55}    & \textbf{90.30}     & \textbf{94.42 (+0.43)}           \\ \hline
CLIP-ViT-B-224/16          & 98.74    & 90.64     & 94.69                   
\end{tabular}
\caption{Linear probe performance of CLIP-ViT-B-224/16 on CIFAR-10 and CIFAR-100 before and after VLP. PT is short for Pre-Training.}
\label{result-cifar}
\end{table*}

\subsection{Main Result}
\subsubsection{Cross-modal Tasks}
We compare UNIMO-3 to a variety of state-of-the-art models on cross-modal, visual and textual task. As shown in Table \ref{result-vqa} and \ref{result-flickr-snlive}, we compare with a series of existing VLP models, including with only 4M images for pre-training models ViLT \cite{kim2021vilt}, UNITER \cite{chen2020uniter}, UNIMO \cite{li2020unimo}, UNIMO-2 \cite{li2022unimo}, ALBEF \cite{li2021align}, VLMo \cite{bao2021vlmo}, METER \cite{dou2022empirical} and BridgeTower \cite{xu2022bridge}, and with a larger number images for pre-training models ALBEF, OFA \cite{wang2022ofa}, SimVLM \cite{wang2021simvlm} and BLIP \cite{li2022blip}. The base-size UNIMO-3 show competitive performances compared with the existing VLP models on downstream VL tasks. The best scores on each metric are marked in bold. UNMIO-3 achieves state-of-the-art performance on VQAv2 dataset, outperforming the preious SOTA model BridgeTower by 0.30\% and 0.49\% on test-dev and test-std. On the SNLI-VE dataset, UNIMO-3 demonstrate a stronger ability to determine the logical relationship between a natural language statement and an image. And on the Flickr30k dataset, UNIMO-3 also have impressive performance in recall metrics, outperforms even some larger-size models.


\begin{figure*}[ht]
    \centering
    \includegraphics[width=14cm]{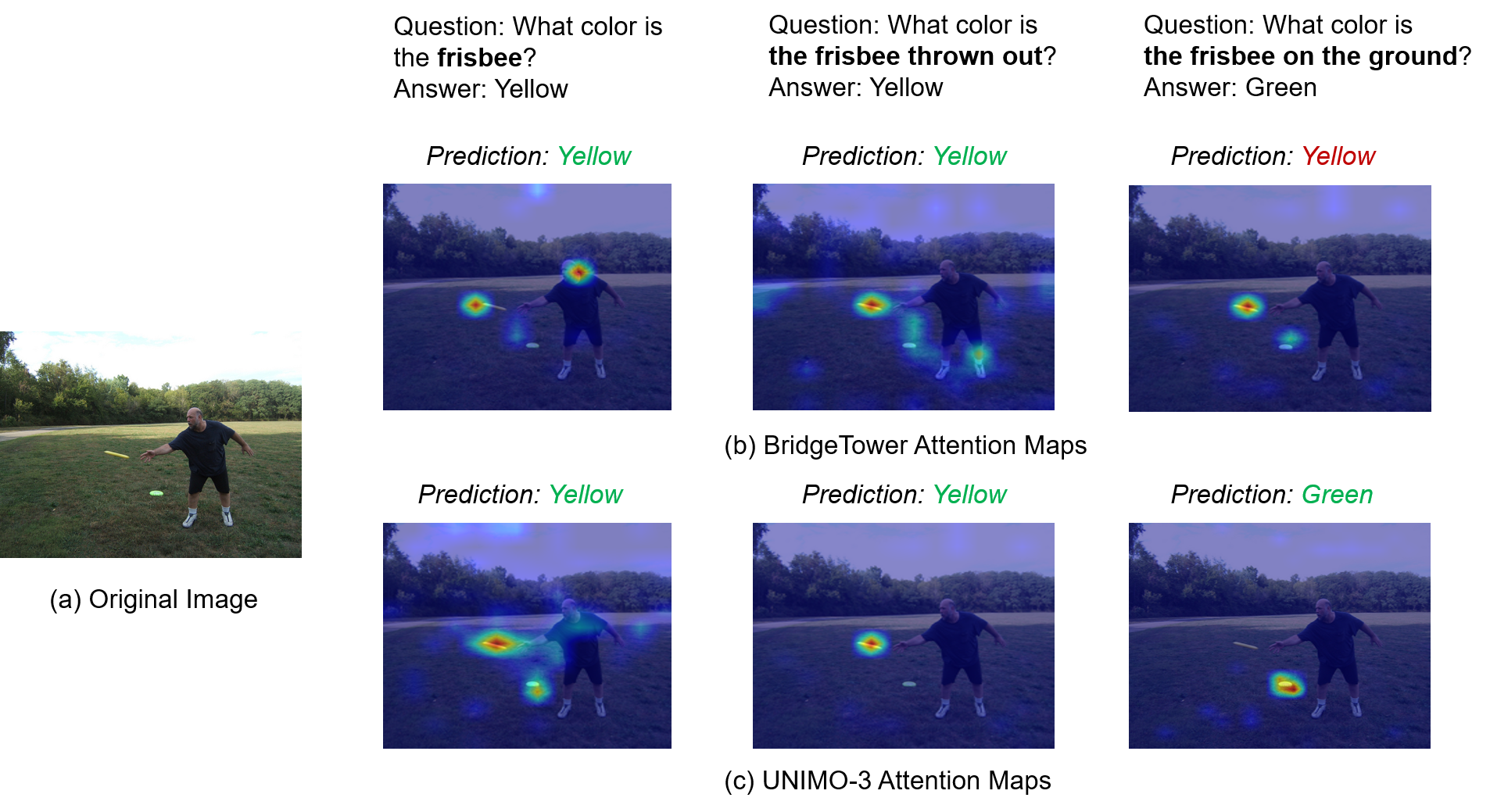}
    \caption{Visualization of the attention maps of BridgeTower and UNIMO-3 models. The example comes from the VQAv2 validation set. Predictions come from the fine-tuning checkpoints of both models.}
    \label{fig-case}
\end{figure*}

\subsubsection{Textual Tasks}
To show the effectiveness of UNIMO-3 on textual tasks, we further compare with both VLP models including METER and BridgeTower, and pre-trained language model RoBERTa. The comparison results in Table \ref{result-glue} demonstrate that UNIMO-3 achieves better performance than existing VLP models including METER and BridgeTower, and achieves comparable performance than existed PLMs such as RoBERTa on GLUE tasks. Expecially, after vision language pre-training, on the RTE task UNIMO-3 textual encoder obtains 0.62 absolute gains compared to RoBERTa-base model.
\subsubsection{Visual Tasks}
For visual tasks, we evaluate UNIMO-3 on CIFAR-10 and CIFAR-100 datasets. Table \ref{result-cifar} shows that, after vision language pre-training, the performance of our visual encoder drops most slightly on both tasks, achieves higher performance compared to METER and BridgeTower, average 0.43\% accuracy improvement than METER. This further proves that the cross-layer and multi-granularity interaction mechanism adopted by the UNIMO-3 model can improve the fine-grained fusion of multimodal information while hardly affecting the effect of the visual encoder.

\subsubsection{Scaling the Model}
Despite the noteworthy results obtained by the UNIMO-3 model in a series of downstream VL tasks, we still expect the cross-layer interaction to show stronger performance on larger-scale models, thus we verified the performance of the scaled-up UNIMO-3-Large model. We replaced the UNIMO-3 uni-modal encoder with the corresponding large versions, we utilize CLIP-ViT-L/14 with 304M parameters for the visual encoder and RoBERTa-Large with 355M parameters for the textual encoder. For each layer of the cross-modal encoder, the hidden size was set to 1,024, the intermediate size of feed-forward networks was set to 4,096, and the number of heads was set to 16. Following the scaling-up version of BridgeTower, we set the patch size to 14x14, the image resolution to 294x294 during pre-training, and the image resolution to 574x574 during fine-tuning on VQAv2. As shown in Table \ref{result-vqa}, UNIMO-3-Large achieves 81.26 accuracy and 81.27 accuracy on the VQAv2 test-dev and test-std set.

\subsection{Visualization}
To demonstrate the effectiveness of the cross-layer multi-granularity interaction, we compare the pre-trained BridgeTower and UNIMO-3 models by analyzing the averaged attention distance \cite{xie2022revealing} of different attention heads in both uni-modal layers and cross-modal layers. The average attention distance is a metric that measures how much other tokens each piece of tokens pays attention to. Similar to the receptive field in CNN that shows how much each pixel depends on other pixels. Different layers have different average attention distances, which indicates that they pay attention to different ranges of information. The higher the average attention distance, the more information each layer pay attention to, and vice versa.

As shown in Figure \ref{fig-case}, we found that (a) In both the uni-modal encoder and cross-modal encoder, we observe a significant difference in average attention distance between different layers. Specifically, the average attention distance at the lower level varies widely across different heads, while it is relatively consistent in the upper level and tends to be higher. This suggests that the cross-layer interaction we apply at higher layers focuses on more granularity information. (b)  In contrast to the BridgeTower model, our model exhibits a higher value of the overall average attention distance. This implies that our model can focus on more granularity multimodal interaction at each layer and achieve more effective cross-modal alignment and fusion.

\section{Conclusion}
In this paper, we propose the UNIMO-3 model with stronger modeling ability for cross-modal fine-grained interaction. UNIMO-3 employs gating mechanisms to adaptively construct connections between each layer of the uni-modal encoder and each layer of the cross-modal encoder. Cross-layer interaction of different modal features enables effective interactive fusion of text and visual semantic information with different granularity. Sufficient experiments have proved that the UNIMO-3 model can achieve impressive performances in a series of downstream tasks.



\section*{Ethics Statement}
Our work complies with ACL Ethics, and all the codes and datasets used in our work comply with the ethics policy.


\bibliography{custom}
\bibliographystyle{acl_natbib}

\appendix



\end{document}